\documentclass[11pt,a4paper]{article}
\usepackage[hyperref]{naaclhlt2018}
\usepackage{times}
\usepackage{latexsym}
\usepackage{graphicx}
\usepackage{hyperref}
\usepackage{url}
\usepackage{pbox}

\aclfinalcopy 

\author{Anton Osika, Susanna Nilsson, Andrii Sydorchuk, Faruk Sahin, Anders Huss \\ 
  Sana Labs, Nybrogatan 8, 114 34 Stockholm, Sweden \\ 
  {\tt \{anton, susanna, andrii, faruk, anders\}@sanalabs.com} \\}

\date{}

\title{Second Language Acquisition Modeling: An Ensemble Approach}

\date{}

\begin{document}
\maketitle
\begin{abstract}

Accurate prediction of students’ knowledge is a fundamental building block of personalized learning systems. Here, we propose a novel ensemble model to predict student knowledge gaps. Applying our approach to student trace data from the online educational platform Duolingo we achieved highest score  on both evaluation metrics for all three datasets in the 2018 Shared Task on Second Language Acquisition Modeling. We describe our model and discuss relevance of the task compared to how it would be setup in a production environment for personalized education.

\end{abstract}

\section{Introduction}
Understanding how students learn over time holds the key to unlock the full potential of adaptive learning. Indeed, personalizing the learning experience, so that educational content is recommended based on individual need in real time, promises to continuously stimulate motivation and the learning process \citep{bauman}. Accurate detection of students' knowledge gaps is a fundamental building block of personalized learning systems \citep{knowledge_gaps} \citep{lindsey2014improving}. A number of approaches exists for modeling student knowledge and predicting student performance on future exercises including IRT \citep{irt}, BKT \citep{sequencing} and  DKT \citep{dkt}. Here we propose an ensemble approach to predict student knowledge gaps which achieved highest score on both evaluation metrics for all three datasets in the 2018 Shared Task on Second Language Acquisition Modeling (SLAM) \cite{slam18}. We analyze in what cases our models' predictions could be improved and discuss the relevance of the task setup for real-time delivery of personalized content within an educational setting.

\section{Data and Evaluation Setup}
The 2018 Shared Task on SLAM provides student trace data from users on the online educational platform Duolingo \citep{slam18}. Three different datasets are given representing user’s responses to exercises completed over the first 30 days of learning English, French and Spanish as a second language. 
Common for all exercises is that the user responds with a sentence in the language learnt. Importantly, the raw input sentence from the user is not available but instead the best matching sentence among a set of correct answer sentences. The prediction task is to predict the word-level mistakes made by the user, given the best matching sentence and a number of additional features provided. The matching between user response and correct sentence was derived by the finite-state transducer method \citep{fst}.

All datasets were pre-partitioned into training, development and test subsets, where approximately the last 10 \% of the events for each user is used for testing and the last 10 \% of the remaining events used for development . Target labels for token level mistakes are provided for the training and development set but not for the test set. Aggregated metrics for the test set were obtained by submitting predictions to an evaluation server provided by Duolingo. The performance for this binary classification task is measured by area under the ROC curve (AUC) and F1-score. 

Although the dataset provided represents real user interactions on the Duolingo platform, the model evaluation setup does not represent a realistic scenario where the predictive modelling would be used for personalizing the content presented to a user. The reason for this is threefold: Firstly, predictions are made given the best matching correct sentence which would not be known prior to the user answering the question for questions that have multiple correct answers. Secondly, there are a number of variables available at each point in time which represent information from the future creating a form of data leakage. Finally, the fact that interactions from each student span all data partitions means that we can always train on the same users that the model is evaluated for and hence there are never first time users, where we would need to infer student mistakes solely from sequential behaviour. To estimate prediction performance in an educational production setting where next-step recommendations must be inferred from past observations, the evaluation procedure would have to be adjusted accordingly.

\section{Method}
To predict word-level mistakes we build an ensemble model which combines the predictions from a Gradient Boosted Decision Tree (GBDT) and a recurrent neural network model (RNN). Our reasoning behind this approach lies in the observation that RNNs have been shown to achieve good results for sequential prediction tasks \citep{dkt} whereas GBDTs have consistently achieved state of the art results on various benchmarks for tabular data \cite{gbdt_benchmark}. 
Even though the data in this case is fundamentally sequential, the number of features and the fact that interactions for each user are available during training make us expect that both models will generate accurate predictions.
Details of our model implementations are given below.

\subsection{The Recurrent Neural Network}
The recurrent neural network model that we use is a generalisation of the model introduced by Piech \shortcite{dkt}, based on the popular LSTM architecture, with the following key modifications:

\begin{itemize}
\item All available categorical and numerical features are fed as input to the network and at multiple input points in the graph of the network (see \ref{sec:supp_model_details})
\item The network operates on a word level, where words from different sentences are concatenated to form a single sequence
\item Information is propagated backward (as well as forward) in time, making it possible to predict the correctness of a word given all the surrounding words within the sentence
\item Multiple ordinary- as well as recurrent layers are stacked, with the information from each level cascaded through skip-connections \cite{bishop} to form the final prediction
\end{itemize}

In model training, subsequences of up to 256 interactions are sampled from each user history in the train dataset, and only the second half of each subsequence is included in the loss function. The binary target variable representing word-level mistakes is expanded to a categorical variable and set to \textit{unknown} for the second half of each subsequence in order to match the evaluation setup.

Log loss of predictions for each subsequence is minimised using adaptive moment estimation \citep{adam} with a batch size of 32. Regularisation with dropout \citep{dropout} and L2 regularisation \citep{Schmidhuber14} is used for embeddings, recurrent and feed forward layers. Data points are used once over each of 80 epochs, and performance continuously evaluated on 70 \% of the dev data after each epoch. The model with highest performance over all epochs is then selected after training has finished. Finally, Gaussian Process Bandit Optimization \citep{GPBandits} is used to tune the hyperparameters learning rate, number of units in each layer, dropout probability and L2 coefficients.

\subsection{The Gradient Boosted Decision Tree}
\label{method:gbdt}
The decision tree model is built using the LightGBM framework \citep{lgbm} which implements a way of optimally partitioning categorical features, leaf-wise tree growth, as well as histogram binning for continuous variables \citep{lgbm:features}. In addition to the variables provided in the student trace data we engineer a number of features which we anticipate should have relevance for predicting the word level mistakes

\begin{itemize}
\item How many times the current token has been practiced
\item Time since token was last seen
\item Position index of token within the best matching sentence 
\item The total number of tokens in the best matching sentence
\item Position index of exercise within session
\item Preceding token
\item A unique identifier of the best matching sentence as a proxy for exercise id
\end{itemize}

Optimal model parameters are learned through a grid search by training the model on the training set and evaluating on the development set to optimize AUC. The optimal GBDT parameter settings for each dataset can be found in the Supplementary Material \ref{sec:hyperparams}.

\subsection{Ensemble Approach}
\begin{figure}[h]
    \centering
    \includegraphics[width=0.45\textwidth]{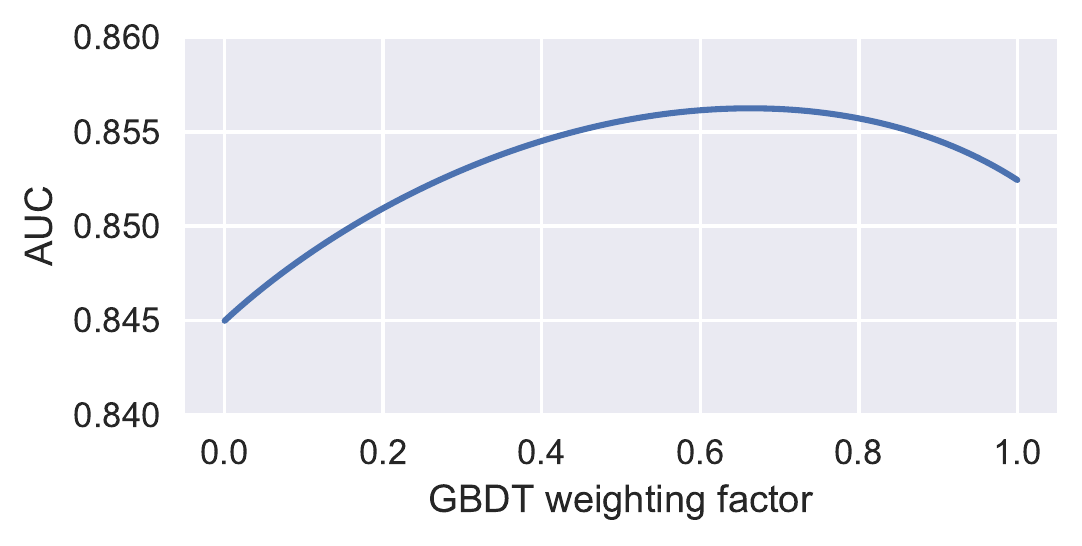}
    \caption{Ensemble model performance as a function of the GBDT ensemble weight parameter for the en\_es dataset.  0.0 is equivalent to using only the neural network model while 1.0 is equivalent to using only GBDT.
}
    \label{fig:ensamble:en_es}
\end{figure}

The predictions generated by the recurrent neural network model and the GBDT model are combined through a weighted average. We train each model using its optimal hyperparameter setting on the train dataset and generate predictions on the dev set. The optimal ensemble weights are then found by varying the proportion of each model prediction and choosing the weight combination which yields optimal AUC score (Figure \ref{fig:ensamble:en_es}).\\

Finally, the RNN and GBDT were trained using their respective optimal hyperparameter settings on the training and development datasets to generate predictions on the test sets. The individual model test set predictions were then combined using the optimal ensemble weights to generate the final test set predictions for task submission.

\section{Discussion}
Our ensemble approach yielded superior prediction performance on the test set compared to the individual performances of the ensemble components (Table \ref{tab_perf_auc}). The F1 scores of our ensemble are reported in Table \ref{tab_perf_f1}.  We note that although the within-ensemble prediction correlations are high (Table \ref{tab_corr}), the prediction diversity evidently suffices for the ensemble combination to outperform the underlying models. This suggests that the RNN and the GBDT differ in performance on different word mistakes. Most likely, the temporal dynamics modelled by the neural network model complement the  GBDT predictions enabling the ensemble to generalise better to unseen user events than its component parts. Notably, none of our individual models would have yielded first place in the Shared Task.

\begin{table}[h]
\centering
\small
\begin{tabular}{c}
\begin{tabular}{|l|l|l|l|}
\hline
Model  &\verb|fr_en|& \verb|es_en| & \verb|en_es|\\\hline
{RNN} & 0.841 & 0.830 & 0.851 \\\hline
{GBDT} & 0.853 & 0.836  &  0.856\\\hline
{Ensemble} & 0.857 & 0.838  &  0.861\\

\hline
\end{tabular}
\end{tabular}
\caption{\label{tab_perf_auc} Model AUC scores on the test partition for all datasets.}
\end{table}

\begin{table}[h]
\centering
\small
\begin{tabular}{c}
\begin{tabular}{|l|l|l|l|}
\hline
  &\verb|fr_en|& \verb|es_en| & \verb|en_es|\\\hline
{Ensemble} & 0.573 & 0.530  &  0.561\\

\hline
\end{tabular}
\end{tabular}
\caption{\label{tab_perf_f1} Model F1 scores on the test partition for all datasets.}
\end{table}

\begin{table}[h]
\centering
\small
\begin{tabular}{c}
\begin{tabular}{|l|l|l|l|}
\hline
Data partition  &\verb|fr_en|& \verb|es_en| & \verb|en_es|\\\hline
{dev} & 0.881 &  0.901& 0.896 \\\hline
{test} & 0.884 & 0.894  &  0.898\\\hline

\end{tabular}
\end{tabular}
\caption{\label{tab_corr} Pearson correlations coefficients between the GBDT and RNN predictions on the dev and test set for all datasets.}
\end{table}

\subsection{Feature Importance}
Given the predictive power of our model we can use the model components to gain insight into what features are most valuable when inferring student mistake patterns. When ranking GBDT features by information gain, we note that 4 out of 5 features overlap between the three datasets (Figure \ref{tab_feature_importance}). The unique user identifier is ranked as second on all datasets, suggesting that very often a separate subtree can be built for each user. This implies that generalisation to new users for the GBDT model would result in performance degradation.


\begin{table}[h]
\centering
\begin{tabular}{c}
\begin{tabular}{|l|l|l|}
\hline
\verb|fr_en|& \verb|es_en| & \verb|en_es|\\ \hline
\textit{token} &  \textit{token} & \textit{token} \\
\textit{user} &  \textit{user} & \textit{user} \\
\textit{format} &  \textit{format} & \textit{format} \\
\textit{exercise id} &  \textit{exercise id} & \textit{exercise id} \\
\textit{time} &  \textit{token attempt} & \textit{time} \\ \hline

\end{tabular}
\end{tabular}
\caption{\label{tab_feature_importance} The top 5 GBDT model features by information gain.
}
\end{table}

\subsection{Relevance for Real Time Prediction Delivery
}


In the setup at hand we have a unique identifier and most of the data available for each user during model training. 
This means that for example the GBDT can naturally build a subtree representing each individual user. For the model evaluation setup where there is no need to generalize to new users this is not an issue.
In a production setting however, the model has to serve new users, which would then have to be handled separately.
Frequent retraining of the model would also be necessary to prevent performance degradation.
This means that the unique user identifier is typically replaced by engineered features that represent the user history. An alternative would be to apply state based models such as Recurrent Neural Networks which by default encode user history without computational overhead or extra engineering effort.

\subsection{Error Analysis}
Although the predictive power of our model is high, there are mistake patterns that our model is not able to capture. The following sections cover two ways of characterizing subsets of the data where the model performs worse than on average. These observations could potentially be used to improve the overall model performance. 

\subsubsection{Performance Decay over Time}
\begin{figure}[h]
    \centering
    \includegraphics[width=0.45\textwidth]{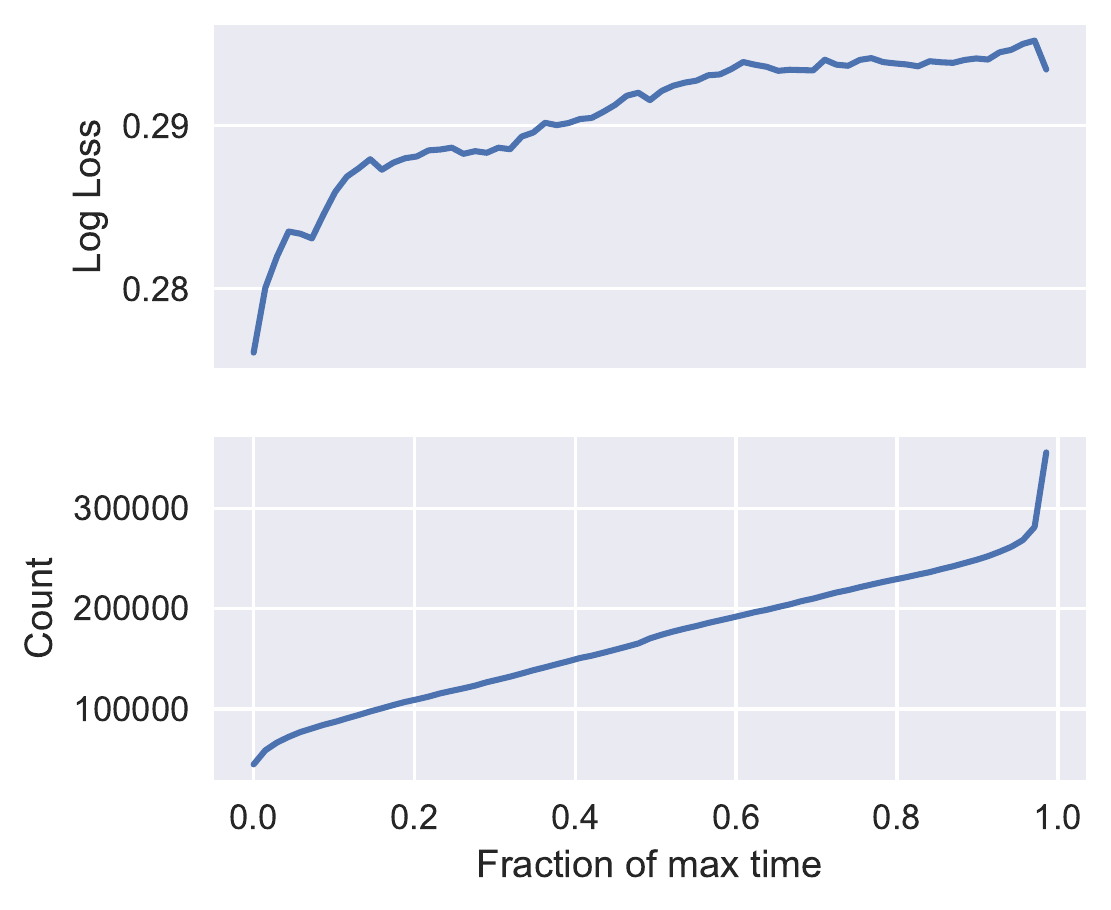}
    \caption{
    Performance decays as instances further away from the label horizon are considered. Log Loss is computed considering only instances before a given fraction of time, where time is normalized by the maximum time for each user. Here performance decay for the en\_es dataset.
}
    \label{fig:decay}
\end{figure}

Due to the sequential partitioning of the training, development and test subsets, the model does not have information about each user's mistakes for the most recent events. In Figure \ref{fig:decay} we note that this lack of information results in a degradation in performance as the predictions get further away from the horizon of labeled data points. Effects which drive this phenomenon include:

\begin{enumerate}
\item The data is non-stationary, i.e. the distribution it comes from varies over time 
\item The model has seen less relevant information about each user when the prediction  is far away from the label horizon
\item  \label{overconfident} The model is overconfident far away from the label horizon since it has never experienced missing information on a user level during training
\end{enumerate}

We note that \ref{overconfident} would not be an issue if the model setup did not include a unique user identifier, which would be desirable in a production setting. For models that do include a unique user identifier as a feature, one way to potentially overcome this performance degradation would be to systematically sample subsequences of the training dataset on a user level, train models separately for each sample and then combine the models. In this way each submodel should be less reliant on the most recent exercise answers at any point in time and thus generalise better to the evaluation setup. This is in effect bagging with a sampling strategy taking consecutive time steps into account \citep{bagging}. We did not attempt to apply this error correction here but leave it for future work.

\subsubsection{The Influence of Rare Words}
\begin{figure}[h]
    \centering
    \includegraphics[width=0.45\textwidth]{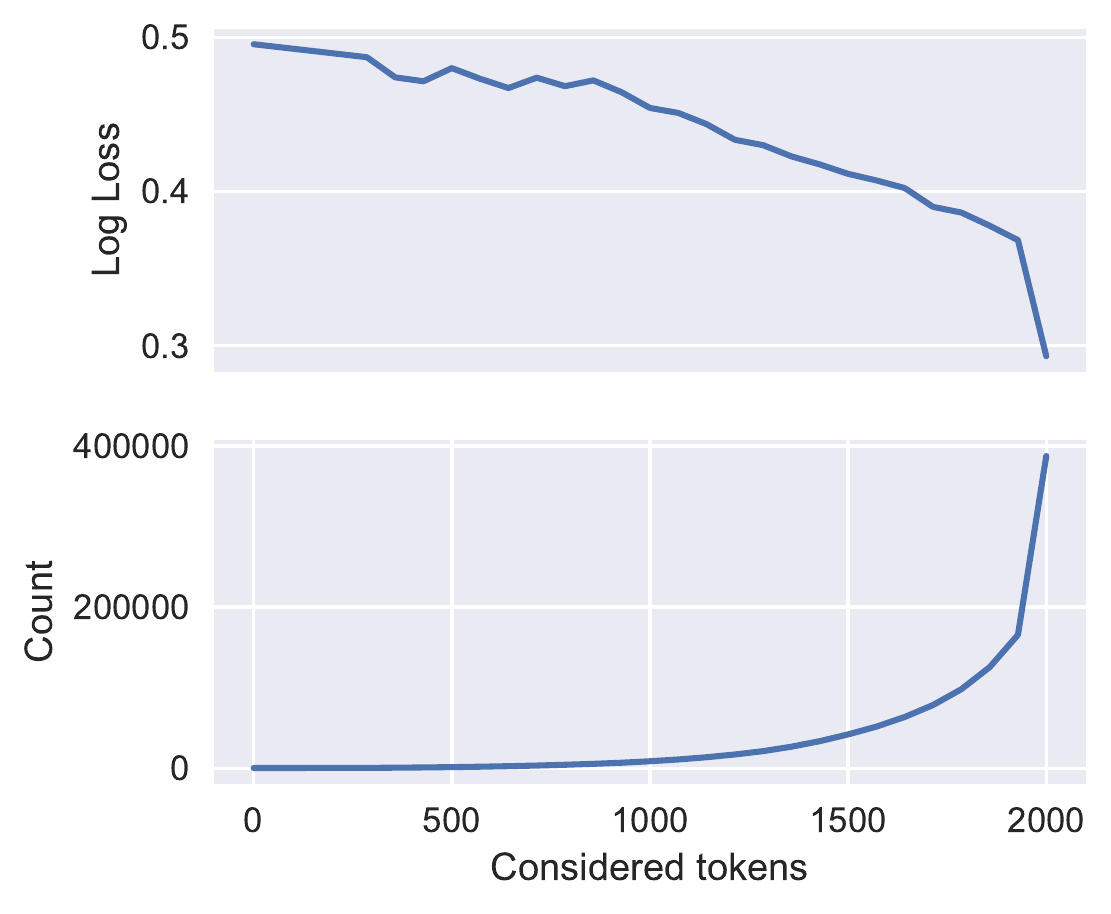}
    \caption{
	Log loss is high when considering only the $x$ most rare tokens and low when considering all tokens on the en\_es dev partition.
}
    \label{fig:rare_words}
\end{figure}

We note that the 4\% of instances with the least common words contribute to 10\% of the prediction error measured in Log Loss, Figure \ref{fig:rare_words}.
This insight gives opportunity to increase prediction performance. Although not attempted here, future work includes building another ensemble component specialized in predicting mistake patterns of words not previously encountered.\\

In conclusion, we have developed an ensemble approach to modeling knowledge gaps applied here within a second language acquisition setting. Albeit not evaluated in a realistic production environment, our ensemble model achieves high predictive performance and allows insights about student mistake patterns. Thus our approach provides a foundation for further research on knowledge acquisition modeling applicable to any educational domain.


\bibliography{naaclhlt2018}
\bibliographystyle{acl_natbib}


\newpage

\appendix

\section{Supplemental Material}
\label{sec:supplemental}

\subsection{The recurrent neural network model design}
\label{sec:supp_model_details}
Our neural network model desisgn is described below:
\begin{enumerate}
\item For each word the network takes as input all available categorical features, excluding morphological features for each word. The exclusion was motivated by the fact that predictive ability added by morphological features was low when evaluated by a  decision tree model.

\item Preprocessed numerical features for \textit{days} and \textit{time} are concatenated to an input vector. (Preprocessing in this case means to normalize to mean zero, variance 1, remove outliers that are larger than 100, and concatenate the value itself with the value exponentiated to 0.5 as well as 2.0)
\item The categories \textit{token}, \textit{part\_of\_speech}, \textit{format}, \textit{correct} and \textit{exercise id} (as described in \ref{method:gbdt}), are each mapped to an embedding vector of length 15.
\item The above categorical features are further combined with the feature \textit{correct} by using the cartesian product, and then mapping each category to an embedding vector.
\item All categorical embeddings and numerical features are concatenated together forming an input vector.
\item The input vector is fed through a two layer bidirectional recurrent neural network, where the input to both of the layers are summed with the output, forming a user state vector.
\item Another input vector is formed by concatenating categorical embeddings for the features \textit{token}, \textit{part\_of\_speech}, \textit{format}, \textit{dependency\_label}, \textit{dependency\_token}, \textit{user\_id} as well as preprocessed numerical features.
\item The user state vector is then projected to two scalars. This is done by dot multiplying it with a vector of trainable variables, as well as dot multiplying it with the second input vector from step 7. The second part accounts for the original operation done by \citep{dkt}.
\item We furthermore compute one scalar for each categorical feature, that is specific for the category of the feature, similar to a logistic regression model.
\item Finally, the second input vector together with all computed scalars are concatenated and fed to a 3 layer feed forward neural network.
\item The sum of all scalar values and the output of the feed forward network forms our logit, which is fed through a sigmoid function outputting the probability of a token level mistake.
\end{enumerate}

\subsection{GBDT Hyperparameters}
\label{sec:hyperparams}

\begin{table}[!ht]

\small
\centering


\begin{tabular}{|l|l|l|l|}

\hline
Model parameter  & fr\_en & es\_en & en\_es\\\hline
num\_leaves & 2400 &  2700 & 2400 \\\hline
n\_estimators & 5744 & 2518  & 3203 \\\hline
learning\_rate & 0.002 & 0.005  &  0.005\\\hline
feature\_fraction & 0.5 & 0.45  &  0.4\\\hline
early\_stopping\_round & 300 & 100  &  100\\
\hline
\end{tabular}
\caption{\label{tab_gbm_pars} Optimal GBDT parameters for all three datasets.}
\end{table}



\end{document}